\documentclass{article}

\usepackage{PRIMEarxiv}

\usepackage[utf8]{inputenc} 
\usepackage[T1]{fontenc}    
\usepackage{hyperref}       
\usepackage{url}            
\usepackage{booktabs}       
\usepackage{amsfonts}       
\usepackage{nicefrac}       
\usepackage{microtype}      
\usepackage{lipsum}
\usepackage{fancyhdr}       
\usepackage{graphicx}       
\usepackage{caption}  
\usepackage{float}    
\usepackage{booktabs}
\usepackage{amsmath, amsthm, amssymb}
\usepackage{amssymb} 
\usepackage{morefloats} 

\newtheorem{theorem}{Theorem}
\newtheorem{definition}{Definition}
\graphicspath{{media/}}     

\title{AFD-STA: Adaptive Filtering Denoising with Spatiotemporal Attention for Chaotic System Prediction
}

\author{
    Chunlin Gong$^{1,4}$, 
    Yin Wang$^{2,3}$\thanks{Corresponding author: \href{mailto:sduwangyin@163.com}{sduwangyin@163.com}},  
    Jingru Li$^5$, 
    Hanleran Zhang$^6$
  \\
  \\
    $^1$ University of Minnesota Twin Cities, Minneapolis, United States \\
    $^2$ School of Statistics and Mathematics, Shandong University of Finance and Economics, Jinan, Shandong, China \\
    $^3$ Jinan Fengdi Intelligent Electronics Co., Ltd., Jinan, Shandong, China \\
    $^4$ Shandong Zhike Intelligence Computing Co., Ltd., Jinan, Shandong, China \\
    $^5$ School of Mechanical,Electrical and Information Engineering, Shandong University, Weihai,Shandong,China \\
    $^6$ SWUFE-UDInstitute of Data Science, Southwestern University of Finance and Economics, Chengdu, Sichuan, China
}

\begin{document}
\maketitle
\begin{abstract}
This paper presents AFD-STA Net, a neural framework integrating adaptive filtering and spatiotemporal dynamics learning for predicting high-dimensional chaotic systems governed by partial differential equations. The architecture combines: 1) An adaptive exponential smoothing module with position-aware decay coefficients for robust attractor reconstruction, 2) Parallel attention mechanisms capturing cross-temporal and spatial dependencies, 3) Dynamic gated fusion of multiscale features, and 4) Deep projection networks with dimension-scaling capabilities. Numerical experiments on nonlinear PDE systems demonstrate the model's effectiveness in maintaining prediction accuracy under both smooth and strongly chaotic regimes while exhibiting noise tolerance through adaptive filtering. Component ablation studies confirm critical contributions from each module, particularly highlighting the essential role of spatiotemporal attention in learning complex dynamical interactions. The framework shows promising potential for real-world applications requiring simultaneous handling of measurement uncertainties and high-dimensional nonlinear dynamics.
\end{abstract}


\section{Introduction}
The translation of natural laws into theoretically grounded frameworks for engineering applications remains a core proposition in interdisciplinary research. Notably, even within deterministic systems, the sensitivity to initial conditions induced by strong nonlinear interactions leads to unpredictability in long-term evolutionary behaviors. Such dynamical systems that exhibit both determinism and unpredictability are collectively termed chaotic systems~\cite{schuster2006deterministic}.

Research on chaotic systems has consistently attracted scholarly attention. Lorenz~\cite{lorenz2017deterministic} demonstrated that minuscule variations in initial conditions could engender substantial divergence in outcomes. This stochastic variability and sensitivity to initial conditions established foundational understanding of chaotic systems. Ruelle and Takens~\cite{ruelle1995turbulence} subsequently proposed the theory of strange attractors, characterizing chaotic systems through fractal dimensions and aperiodic orbits. Chen~\cite{chen1999yet} extended the topological structure of Lorenz attractors, establishing the Chen chaotic model with more complex dynamical behaviors. At the partial differential equation level, the Kuramoto-Sivashinsky equation~\cite{ashinsky1988nonlinear} successfully describes turbulent oscillatory dynamics in reaction-diffusion systems through spatiotemporal chaotic mechanisms. The Swift-Hohenberg equation~\cite{swift1977hydrodynamic} provides an effective chaotic modeling framework for pattern formation in crystal growth processes. The Brusselator equation~\cite{prigogine1968symmetry} offers a robust framework for investigating self-organization and nonlinear phenomena in chemical reaction dynamics through its description of chemical oscillatory behaviors in reaction-diffusion processes.

Chaotic system prediction research has evolved two principal approaches: physics-based numerical methods and data-driven machine learning techniques. The former establishes explicit mathematical representations through differential equation systems~\cite{chen1999yet,ashinsky1988nonlinear,swift1977hydrodynamic,prigogine1968symmetry}, solved via numerical discretization techniques such as spectral methods~\cite{canuto2007spectral} and finite difference methods~\cite{press2007numerical}. However, computational complexity typically escalates exponentially with dimensionality, and the construction of precise mathematical models heavily relies on domain-specific prior knowledge. The latter employs machine learning architectures like deep neural networks~\cite{karniadakis2021physics} to directly learn implicit evolutionary patterns from high-dimensional observational data, significantly reducing dependence on complete mechanistic models.

Current research presents numerous neural network applications for chaotic system prediction. Paper~\cite{cheng2021high} proposed a temporal convolutional network incorporating spatial and channel attention mechanisms via convolutional block attention modules for chaotic systems and solar activity time series prediction. Study~\cite{chen2021chaotic} developed an echo state network optimized through selective opposition grey wolf optimization for input weight matrix reconstruction and enhanced search capability, applied to Mackey-Glass and Lorenz chaotic system prediction. Work~\cite{dudukcu2023temporal} introduced a hybrid architecture combining temporal convolutional networks with recurrent neural network layers for Lorenz system and arrhythmia ECG time series prediction. Research~\cite{nasiri2022mfrfnn} presented a multifunctional recurrent fuzzy neural network integrating dual TSK fuzzy neural networks with feedback loops and particle swarm optimization, implemented in chaotic systems and industrial applications like gas furnace time series prediction. Investigation~\cite{karasu2022crude} combined long short-term memory networks with technical indicators (trend, momentum, volatility), employing Henry gas solubility optimization with logistic chaotic mapping for feature selection in WTI and Brent crude price prediction.

Our work focuses on data-driven neural network approaches for spatiotemporal chaotic system prediction. We employ phase space reconstruction to establish mapping relationships between delay attractors and original attractors, proposing a novel network architecture: AFD-STA.

\begin{itemize}
\item[\textbf{1)}] Establishing diffeomorphic mapping relationships between delay attractors and original attractors based on Takens' embedding theorem, deriving rigorous mathematical characterizations for spatiotemporal information transformation
\item[\textbf{2)}] Proposing AFD-STA with a spatiotemporal attention gated fusion mechanism for decoupling and reconstructing multi-scale dynamical features
\end{itemize}

\section{Preliminary}
\label{Section 2}
For PDE systems exhibiting complex spatiotemporal chaotic characteristics, the time-delay embedding theorem extracts the implicit manifold structure and establishes a topology-preserving mapping relationship from low-dimensional chaotic attractors to high-dimensional spatiotemporal patterns. This provides theoretical guarantees for embedding dynamical priors in subsequent machine learning models.

Takens embedding theorem~\cite{takens2006detecting} serves as a powerful tool for studying chaotic time series. The theorem states:\\

\begin{theorem}
For an infinite-length, noise-free scalar time series $x(t)$ from a $d$-dimensional chaotic attractor, there exists an $m$-dimensional embedded phase space that preserves topological equivalence when $m \geq 2d+1$.
\label{theorem 1}
\end{theorem}

According to Takens embedding theorem, we can reconstruct the corresponding high-dimensional phase space from a one-dimensional chaotic time series, ensuring topological equivalence with the original system.

Typically, phase space reconstruction employs coordinate delay embedding, constructing phase space vectors through delay time $\tau$ and embedding dimension $m$:
\begin{equation}
X(t) = [x(t), x(t+\tau), x(t+2\tau), \ldots, x(t+(m-1)\tau)]
\label{equation 1}
\end{equation}
where $\tau$ denotes the time delay and $m$ represents the embedding dimension.

The generalized embedding theorem proposed by Deyle et al.~\cite{deyle2011generalized} provides a methodology for attractor reconstruction using multivariate time series. Based on this theory, multivariate phase space vectors can be constructed to analyze intrinsic relationships between spatial variables. The embedding vector for multivariate time series can be defined as:
\begin{equation}
X(t) = [y_1(t), y_2(t), \ldots, y_d(t), \ldots, y_1(t+(m-1)\tau), \ldots, y_d(t+(m-1)\tau)]
\label{equation 2}
\end{equation}

Building upon the generalized embedding theorem, Chen et al.~\cite{chen2020predicting} proposed an innovative method for reconstructing univariate time series attractors from high-dimensional time series:\\

\begin{definition}
Let $O$ denote the observed system attractor with box dimension $d_o$. The time series $X(t_m) = [x_1(t_m), \ldots, x_n(t_m)] \in \mathbb{R}^n$ represents the system state on $O$ at time $t_m$, where $x_i(t_m)$ denotes the $i$-th component at the $m$-th time step. The delay coordinate map $\Psi: \mathbb{R}^n \to \mathbb{R}^{L+1}$ is defined as:
\begin{equation}
\Psi(X(t_m)) = [x_k(t_m), x_k(t_{m+1}), \ldots, x_k(t_{m+L})] = Z(t_m)
\end{equation}
where $x_k$ represents our target variable of interest.
\end{definition}

Combining the Equation (\ref{equation 2}) with \autoref{theorem 1}, we obtain:

\begin{theorem}
Let $O$ be the observed attractor with box dimension $d_o$, and $L$ the future time step length for reconstruction. If $L$ satisfies $L > 2d_O$, the reconstructed attractor $Z(t_m) = [x_k(t_m), x_k(t_{m+1}), \ldots, x_k(t_{m+L})] \in \mathbb{R}^{L+1}$ from time series $X(t_m)$ maintains a topological conjugacy relationship with the original attractor $O$.
\label{theorem 2}
\end{theorem}

For a known $m$-dimensional time series $X(t_m)$ with $M$ sampling points, the following mapping relationships exist:\\
\begin{equation}
\left\{ 
\begin{aligned}
\Psi(X(t_m)) &= Z(t_m), & m &= 1,2,\ldots,M \\
\Phi(Z(t_m)) &= X(t_m), & m &= 1,2,\ldots,M 
\end{aligned}
\right.
\label{eq:mapping}
\end{equation}

Considering that $Z(t_m)$ consists of $L+1$ temporal components of $x_k$, $\Psi$ can be expressed as a family of injective functions $\{\Psi_1, \Psi_2, \ldots, \Psi_{L+1}\}$.

The matrix formulation reveals variable mapping relationships across different time steps:\\
\begin{equation}
\begin{pmatrix}
\Psi_1(X(t_1)) & \Psi_1(X(t_2)) & \cdots & \Psi_1(X(t_M)) \\
\Psi_2(X(t_1)) & \Psi_2(X(t_2)) & \cdots & \Psi_2(X(t_M)) \\
\vdots & \vdots & & \vdots \\
\Psi_{L+1}(X(t_1)) & \Psi_{L+1}(X(t_2)) & \cdots & \Psi_{L+1}(X(t_M))
\end{pmatrix}
= 
\begin{pmatrix}
x_k(t_1) & x_k(t_2) & \cdots & x_k(t_M) \\
x_k(t_2) & x_k(t_3) & \cdots & x_k(t_{M+1}) \\
\vdots & \vdots & & \vdots \\
x_k(t_{L+1}) & x_k(t_{L+2}) & \cdots & x_k(t_{M+L})
\end{pmatrix}
\label{eq:matrix}
\end{equation}

This matrix formulation enables comprehensive capture of nonlinear relationships among multiple time series.

\section{Methodology}

\subsection{AFD-STA Architecture}

PDE systems are difficult to predict due to their spatiotemporal continuity and complex changing relationships. In this paper, we aim to learn system characteristics through discrete sampling points to predict unknown models. Consider an abstract infinite-dimensional system:
\[
\frac{\partial u}{\partial t} = f(u; x).
\]
where $u$ is the unknown variable, and $f$ is an arbitrary operator of any order or an arbitrarily complex linear or nonlinear function of $u$. Define the spatial interval as $\Delta x$, then $x_i = x_0 + i \Delta x, i = 1, \dots, D$. Similarly, we have $t_j = t_0 + j \Delta t, j = 1, 2, \dots$. Define $u_i^j = u(x_i, t_j)$, then:
\[
u_i^j = g(U^0, U^1, \dots, U^j),
\]
where $U^j = (u_1^j, u_2^j, \dots, u_D^j)$. Through spatiotemporal discretization, we obtain the independent variables $u_1^j, u_2^j, \dots, u_D^j$, and the original infinite-dimensional system is transformed into a $D$-dimensional dynamical system. Thus, the above formula can be regarded as the observed values of the original system.

\begin{figure}[H]
  \centering
  \includegraphics[width=6in]{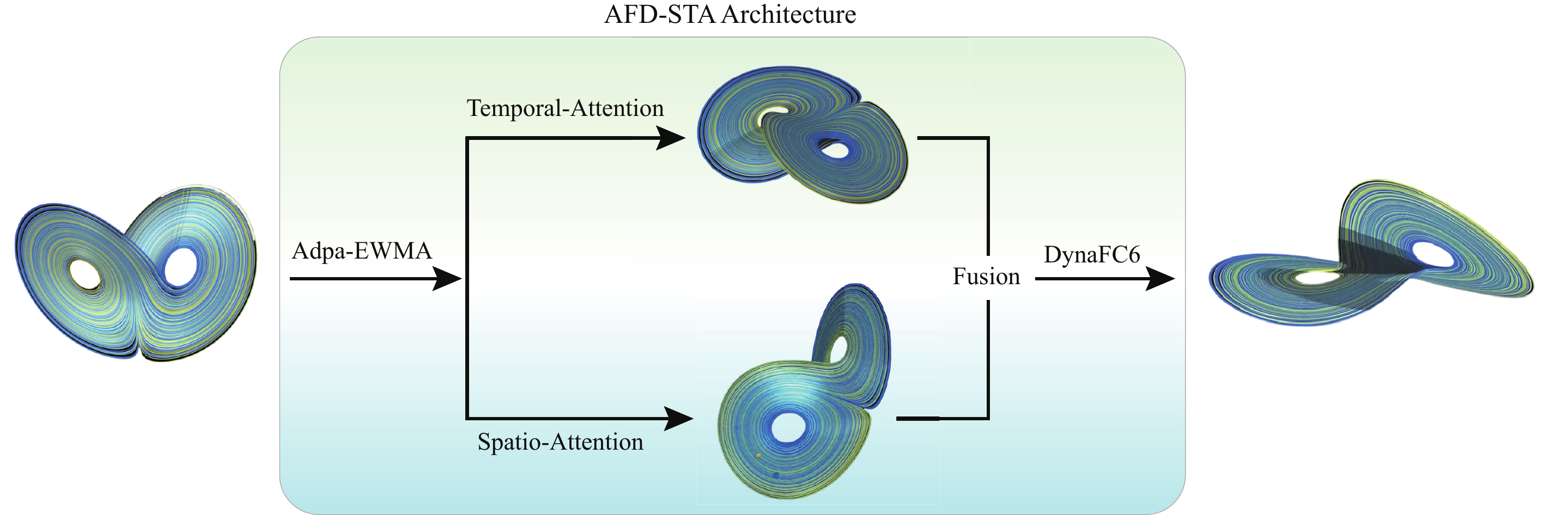} 
  \caption{Demonstrates the overall strcuture.}  
  \label{fig:1}  
\end{figure}

\begin{figure}[H]
  \centering
  \includegraphics[width=6in]{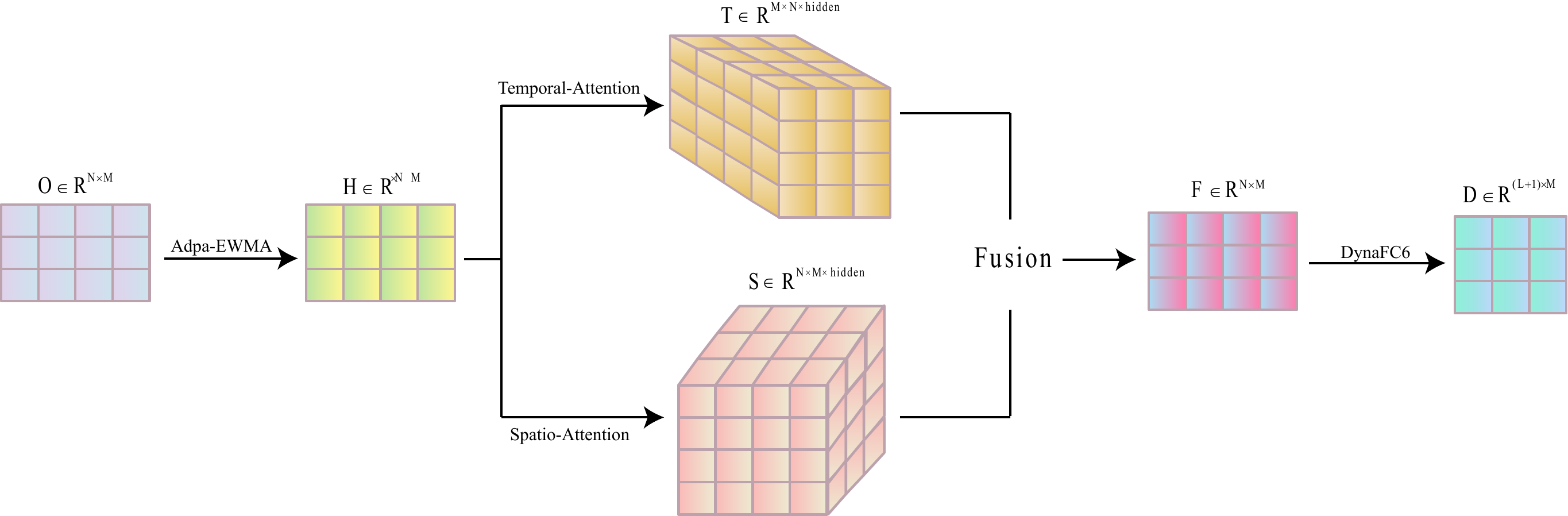} 
  \caption{Demonstrates the the matrix dimension changes under each module’s processing.}  
  \label{fig:2}  
\end{figure}

Consider that our model aims to predict the delayed attractor $D = [x_k^{t_{M+1}}, \dots, x_k^{t_{M+L}}]$ at point $x_k$ for $L$ future steps by observing the original attractor $O = [U^{t_1}, U^{t_2}, \dots, U^{t_M}]^T$ composed of $M$ observations of the original system. From \autoref{theorem 2}, when the dimension $d_O$ of the original attractor $O$ satisfies $L \geq 2d_O + 1$, the delayed attractor $D$ is topologically conjugate to the original attractor $O$. From Equation (\ref{equation 1}), the original attractor and the delayed attractor satisfy:
\[
\begin{cases}
\Psi(O) = D \\
\Phi(D) = O
\end{cases}.
\]
From the \autoref{Section 2}, we can obtain the STI spatiotemporal information transformation equation for the original attractor and delayed attractor, as shown in Equation (\ref{eq:matrix}):
\[
\begin{pmatrix}
\Psi_1(U^{t_1}) & \Psi_1(U^{t_2}) & \cdots & \Psi_1(U^{t_M}) \\
\Psi_2(U^{t_1}) & \Psi_2(U^{t_2}) & \cdots & \Psi_2(U^{t_M}) \\
\vdots & \vdots & & \vdots \\
\Psi_{L+1}(U^{t_1}) & \Psi_{L+1}(U^{t_2}) & \cdots & \Psi_{L+1}(U^{t_M})
\end{pmatrix}
=
\begin{pmatrix}
u_k^{t_1} & u_k^{t_2} & \cdots & u_k^{t_M} \\
u_k^{t_2} & u_k^{t_3} & \cdots & u_k^{t_{M+1}} \\
\vdots & \vdots & & \vdots \\
u_k^{t_{L+1}} & u_k^{t_{L+2}} & \cdots & u_k^{t_{M+L}}
\end{pmatrix}.
\]
Next, we construct a suitable neural network model to fit the nonlinear mapping $\Psi$. As shown in \autoref{fig:1}, we design the AFD-STA Net as a nonlinear fitting neural network. To address the noise impact from the real world, a common solution is to use EWMA for data filtering. Our model introduces adaptive parameters on top of EWMA, proposing the Adap-EWMA module. AFD-STA first applies the Adap-EWMA module to the observed original attractor $O \in \mathbb{R}^{N \times M}$ for data smoothing, obtaining $H \in \mathbb{R}^{N \times M}$. To handle the influence of the remaining time series information of a node on its value at the current time, we use the Temporal-Attention module to expand the features of each time step and extract and fuse features from other time steps via the Self-Attention mechanism, obtaining $T \in \mathbb{R}^{M \times N \times \text{hidden}}$. Meanwhile, to measure the influence of other points in the system on the target point, we use the Spatio-Attention module to expand the features of each spatial point and perform feature extraction and fusion, obtaining $S \in \mathbb{R}^{N \times M \times \text{hidden}}$. The Fusion module employs a gated fusion mechanism to dynamically weight and fuse temporal and spatial feature information, followed by feature compression, yielding $F \in \mathbb{R}^{N \times M}$. Finally, we use a six-layer fully connected network (DynaFC6) to construct a deep mapping from spatiotemporal features to the delayed attractor. This network effectively extracts temporal dynamic features through intermediate layer upscaling-downscaling operations and captures the complex evolution patterns of the attractor using deep nonlinear transformations. In particular, the residual connection design maintains feature representation capability while ensuring training stability. \autoref{fig:2} illustrates the matrix dimension changes under each module’s processing.

\subsection{Adap-EWMA}

The Exponential Weighted Moving Average (EWMA) \cite{genccay2001introduction} refers to a moving average with exponentially decreasing weights, where the weights of data points decrease exponentially over time, with more recent data weighted more heavily, but older data still given some weight. The EWMA method performs well in application scenarios with real-world noise, but its fixed decay parameters and decay rate make it difficult for the model to learn the noise characteristics of the current system. Considering that adjusting the decay coefficients for each position adaptively would require more training data and slower convergence, while fully data-driven decay coefficients for each time step may lead to poor interpretability and unstable results, this section introduces a learnable parameter as the baseline parameter, setting the position encoding as the decay rate. Through the learnable baseline parameter $\beta$ and the decay rate, we can obtain different decay coefficients for each time step. This method extracts the dynamic trends of the original attractor $O$ to achieve data denoising.

Nonlinear systems typically become progressively unstable over time, in which case we rely more on new observation data. Therefore, we stipulate that the position encoding increases, meaning later positions have greater weights. Define the position encoding vector $j = [0, 1, \dots, M-1] \in \mathbb{R}^M$, and introduce the learnable parameter $\beta$, obtaining the scaled intermediate vector $h$:
\[
h = \beta \cdot j \in \mathbb{R}^M.
\]
Through the sigmoid function, each element in $h$ is mapped to a smoothing coefficient vector:
\[
\alpha_j = \sigma(h_j) = \frac{1}{1 + e^{-h_j}} \in (0,1), \quad \forall j \in \{0, \dots, M-1\}.
\]
This yields $\alpha = [\alpha_0, \alpha_1, \dots, \alpha_{M-1}] \in \mathbb{R}^M$. The smoothed data $H$ is initialized to zero and updated iteratively per time step. For the $j$-th time step of the $i$-th point, the current smoothed data $H(i,j)$ is computed by mixing the current observation $O(i,j)$ with the historical smoothed data $H(i,j-1)$:
\[
H(i,j) = \alpha_j \times O(i,j) + (1 - \alpha_j) \times H(i,j-1),
\]
This process can be expanded into an explicit weighted sum form:
\[
H(i,j) = \sum_{k=0}^j \left( \alpha_k \prod_{l=k+1}^j (1 - \alpha_l) \right) O(i,j),
\]
where the weight parameters satisfy the normalization constraint $\sum_{k=0}^j \alpha_k \prod_{l=k+1}^j (1 - \alpha_l) = 1 - \prod_{i=0}^j (1 - \alpha_i)$. When $\alpha_j \equiv \alpha$ is a constant, the formula degenerates into the standard EWMA form:
\[
H(i,j) = \alpha \sum_{k=0}^j (1 - \alpha)^{j-k} O(i,j).
\]
Through this method, we obtain the sequence for the same point $H^i = [H(i,0), H(i,1), \dots, H(i,M-1)]$, and extend the same process to different points to obtain $H = [H^1, H^2, \dots, H^N]^T \in \mathbb{R}^{N \times M}$.

\subsection{Temporal-Attention and Spatio-Attention Module}

\begin{figure}[H]
  \centering
  \includegraphics[width=7in]{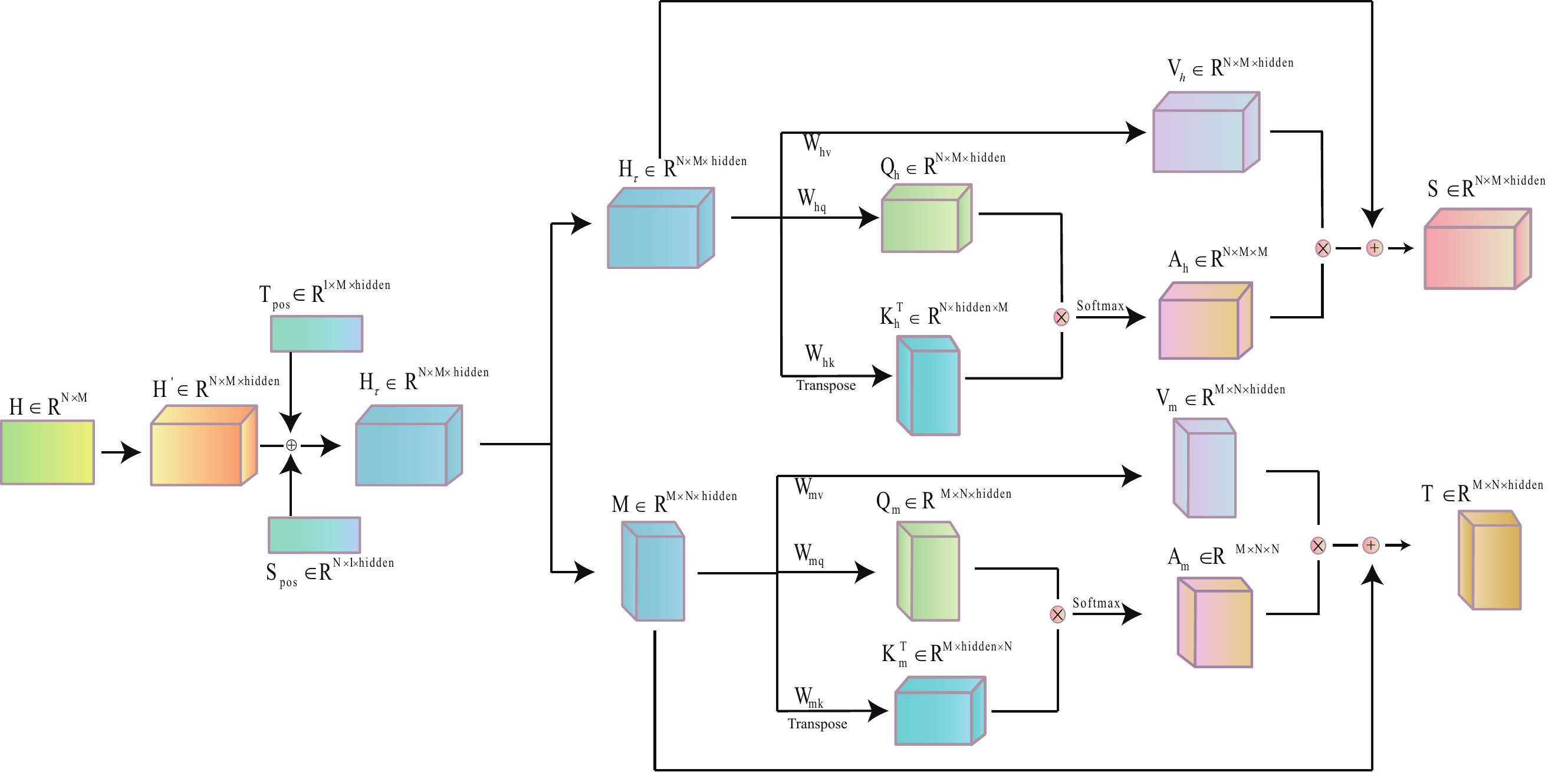} 
  \caption{Spatiotemporal attention architecture}  
  \label{fig:3}  
\end{figure}

Self-Attention \cite{vaswani2017attention} is widely used in LLMs due to its excellent ability to identify correlations within sequences. As shown in \autoref{fig:3}, unlike the conventional application of Self-Attention to our time series, we use parallel temporal and spatial attention mechanisms. This approach not only considers the influence of a point at different times on its value at other times but also accounts for the mutual influence between points at the same time. For the previous step’s $H \in \mathbb{R}^{N \times M}$, we first perform dimensionality expansion on the original matrix, then expand the features of each time step, obtaining the matrix $H' \in \mathbb{R}^{N \times M \times \text{hidden}}$. To account for the impact of different temporal and spatial positions on the model, we introduce randomly initialized, learnable temporal and spatial embedding matrices $T_{\text{pos}} \in \mathbb{R}^{1 \times M \times \text{hidden}}$, $S_{\text{pos}} \in \mathbb{R}^{N \times 1 \times \text{hidden}}$, and obtain $H_\tau$ through broadcasting:
\[
H_\tau = H' \oplus T_{\text{pos}} \oplus S_{\text{pos}} \in \mathbb{R}^{N \times M \times \text{hidden}}.
\]
Considering the Self-Attention calculation of mutual influences between variables in the first dimension, we pass $H_\tau \in \mathbb{R}^{N \times M \times \text{hidden}}$ into spatial attention to compute the mutual influence among $N$ points. Transposing the first and second dimensions of $H_\tau$ yields $M \in \mathbb{R}^{M \times N \times \text{hidden}}$, which is passed into temporal attention to focus on computing mutual influences between time steps. The two matrices are processed through two independent single-head Self-Attention layers to compute:
\[
S = \text{Spatio-Attention}(H_\tau) \in \mathbb{R}^{N \times M \times \text{hidden}},
\]
\[
T = \text{Temporal-Attention}(H_\tau') \in \mathbb{R}^{M \times N \times \text{hidden}}.
\]

\subsection{Fusion Module}

\begin{figure}[H]
  \centering
  \includegraphics[width=7in]{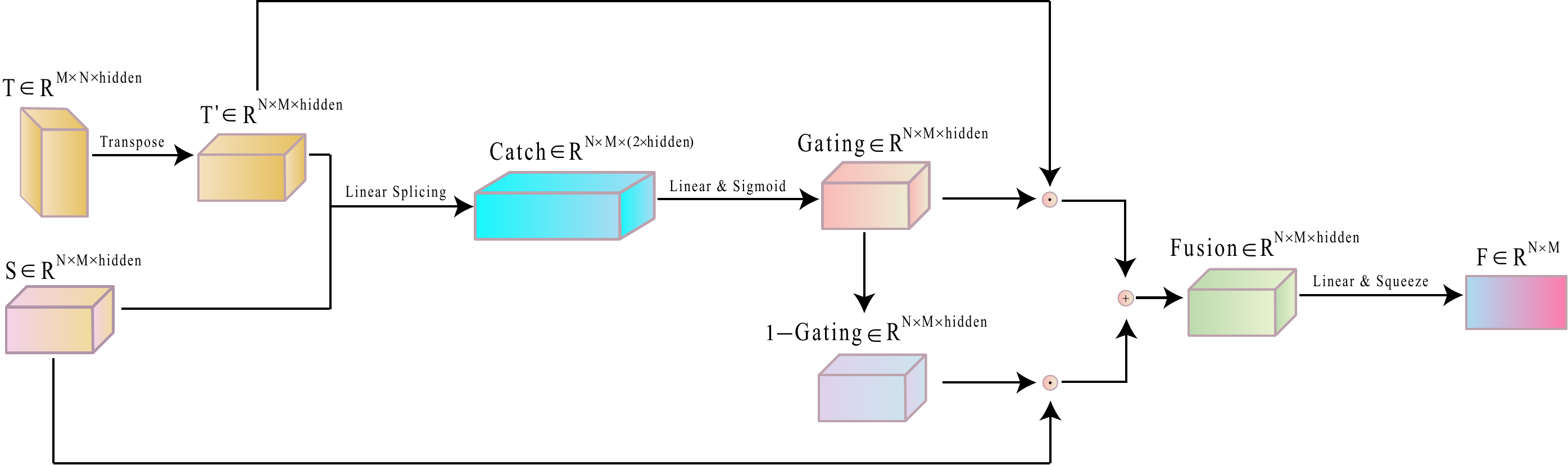} 
  \caption{Fusion module architecture}  
  \label{fig:4}  
\end{figure}

In scenarios involving information aggregation under multiple parallel paths, dynamic weight control methods dominated by gating mechanisms have become mainstream \cite{shazeer2017outrageously}. The overall structure of the gated fusion is shown in \autoref{fig:4}. We first adjust the dimensions of $T \in \mathbb{R}^{M \times N \times \text{hidden}}$ to $T' \in \mathbb{R}^{N \times M \times \text{hidden}}$, and through linear concatenation in the first dimension, we obtain the Catch matrix:
\[
\text{Catch} = [S | T'] \in \mathbb{R}^{N \times M \times (2 \times \text{hidden})},
\]
where the Catch matrix contains both temporal and spatial attention information. To explore the proportion of temporal and spatial attention in the system, we use a linear layer to extract aggregated spatiotemporal information from the Catch matrix, smoothed by the sigmoid function, to obtain the dynamic weight Gating matrix:
\[
\text{Gating} = \text{Sigmoid}(\text{Linear}(\text{Catch}) \in \mathbb{R}^{N \times M \times \text{hidden}}) \in (0,1).
\]
Define the matrix of aggregated spatiotemporal information features as Fusion, computed as:
\[
\text{Fusion} = \text{Gating} \odot T + (1 - \text{Gating}) \odot S,
\]
where $\odot$ denotes element-wise multiplication, and $\text{Fusion} \in \mathbb{R}^{N \times M \times \text{hidden}}$. To aggregate the temporal feature dimension, we apply a linear layer to compress the last dimension’s features into one dimension and perform overall dimension compression to remove the third dimension, obtaining the $F$ matrix:
\[
F = \text{squeeze}_3(\text{Linear}(\text{Fusion}) \in \mathbb{R}^{N \times M \times 1}) \in \mathbb{R}^{N \times M}.
\]

\subsection{DynaFC6 Network}

\begin{figure}[H]
  \centering
  \includegraphics[width=7in]{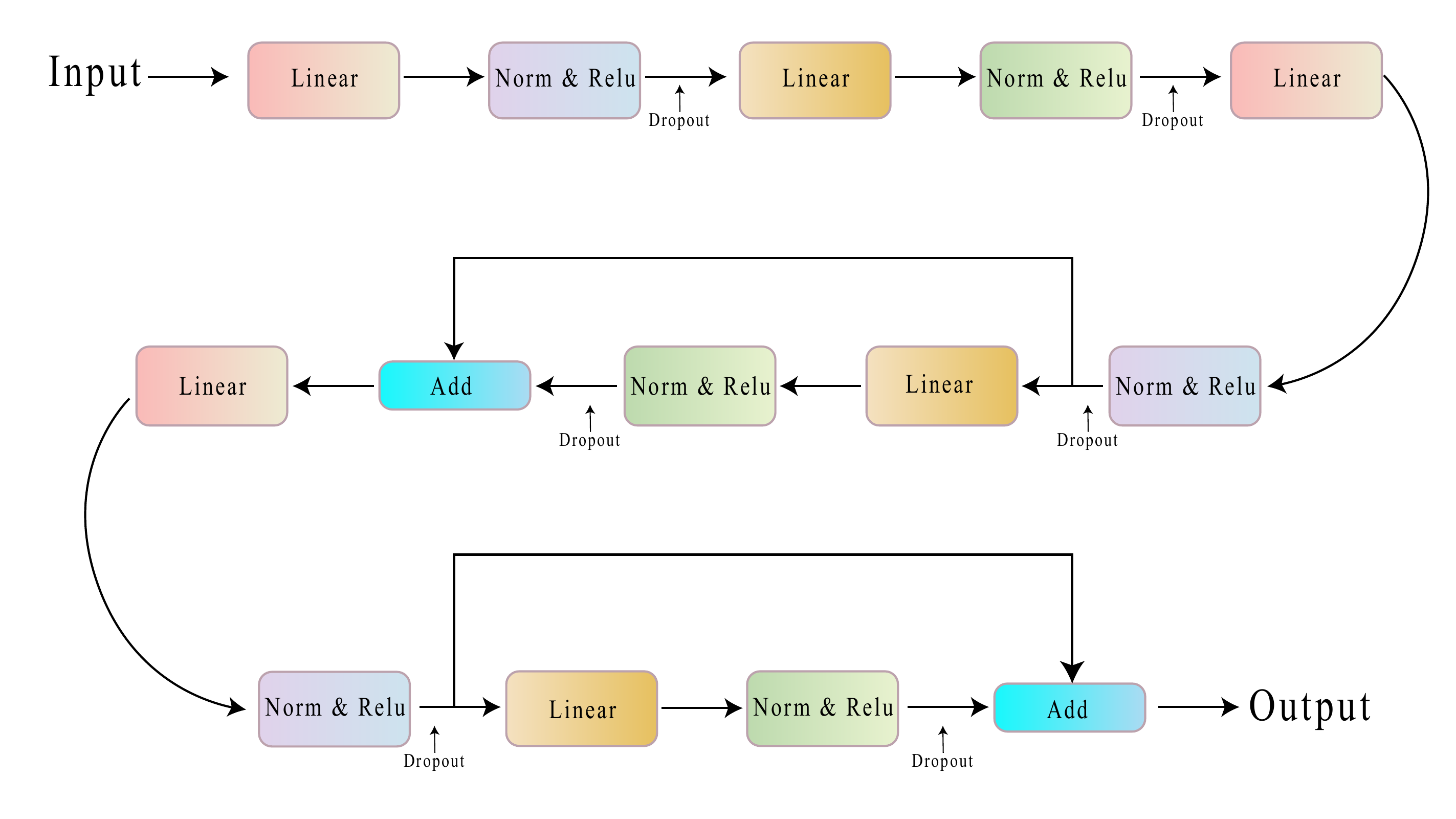} 
  \caption{DynaFC6 Network architecture}  
  \label{fig:5}  
\end{figure}

This module implements the mapping from the spatiotemporal feature matrix $F \in \mathbb{R}^{N \times M}$ to the delayed attractor matrix $D \in \mathbb{R}^{(L+1) \times M}$ through a six-layer fully connected network (as shown in \autoref{fig:5}), with the mathematical form:
\[
D = \Gamma_{\text{out}} \circ \Gamma_6 \circ \cdots \circ \Gamma_1(F^T).
\]
In this section, we use an expansion and compression process. The first two layers expand and deepen the temporal features:
\[
H_1 = \text{Dropout}(\text{ReLU}(\text{LayerNorm}(W_1 F^T + b_1))) \in \mathbb{R}^{M \times \text{hidden}},
\]
\[
H_2 = \text{Dropout}(\text{ReLU}(\text{LayerNorm}(W_2 H_1 + b_2))) \in \mathbb{R}^{M \times \text{hidden}},
\]
In the third layer, we apply feature compression, reducing the hidden layer features to half their size:
\[
H_3 = \text{Dropout}(\text{ReLU}(\text{LayerNorm}(W_3 H_2 + b_3))) \in \mathbb{R}^{M \times \text{hidden}/2}.
\]
The fourth, fifth, and sixth layers maintain the hidden layer feature dimensions, with residual connections \cite{he2016deep} added in the fourth and sixth layers to ensure gradient flow:
\[
H_4 = \text{Dropout}(\text{ReLU}(\text{LayerNorm}(W_4 H_3 + b_4))) + H_3 \in \mathbb{R}^{M \times \text{hidden}/2},
\]
\[
H_5 = \text{Dropout}(\text{ReLU}(\text{LayerNorm}(W_5 H_4 + b_5))) \in \mathbb{R}^{M \times \text{hidden}/2},
\]
\[
H_6 = \text{Dropout}(\text{ReLU}(\text{LayerNorm}(W_6 H_5 + b_6))) + H_5 \in \mathbb{R}^{M \times \text{hidden}/2}.
\]
Finally, through the output layer and transposition, we obtain the delayed attractor matrix:
\[
O = (\text{Linear}(H_6) \in \mathbb{R}^{M \times (L+1)})^T.
\]

\section{Numerical Experiments}

\subsection{Dataset Generation}
\label{Section 4.1}

Consider the nondimensionalized KS equation:
\[
u_t + u_{xx} + u_{xxxx} + u u_x = 0.
\]
Place $N$ equally spaced points in the periodic domain $x \in [1, 16]$. The initial condition is set as:
\[
u(x,0) = \cos\left( \frac{x}{16} \right)
\]
Transform the physical field to the spectral space using the Fast Fourier Transform:
\[
v(k,t) = \mathcal{F}u(x,t) = \frac{1}{N} \sum_{j=1}^N u(x_j, t) e^{-i k x_j}.,
\]
where the wavenumber $k$ satisfies the Nyquist sampling criterion. The linear operator is defined as:
\[
\mathcal{L}(k) = k^2 - k^4.
\]
The fourth-order Exponential Time Differencing scheme ETDRK4 \cite{cox2002exponential} is used to advance the time evolution, with the single-step iteration process as follows:
\[
\begin{array}{l}
a = e^{h \mathcal{L}/2} v_n + h Q \cdot \mathcal{N}(v_n), \\
b = e^{h \mathcal{L}/2} v_n + h Q \cdot \mathcal{N}(a), \\
c = e^{h \mathcal{L}/2} a + h Q \cdot (2 \mathcal{N}(b) - \mathcal{N}(v_n)), \\
v_{n+1} = e^{h \mathcal{L}} v_n + h \left[ f_1 \mathcal{N}(v_n) + 2 f_2 (\mathcal{N}(a) + \mathcal{N}(b)) + f_3 \mathcal{N}(c)\right],
\end{array}
\]
where the time step $h = \Delta t = 0.1$, and the nonlinear term is computed by decoupling in the physical space:
\[
\mathcal{N}(v) = -0.5 i k \cdot \mathcal{F}\left\{ \left[ \mathcal{F}^{-1}\{v\} \right]^2 \right\}.
\]
The key coefficients $Q, f_i$ are calculated via contour integration in the complex plane to avoid numerical singularities:
\[
Q = h \cdot \text{Re}\left( \frac{1}{M_1} \sum_r \frac{e^{h \mathcal{L}/2} - 1}{h \mathcal{L} + r} \right), \quad f_i = h \cdot \text{Re}\left( \frac{1}{M_1} \sum_r \frac{P_i(r)}{(h \mathcal{L} + r)^3} \right). \quad (i=1,2,3)
\]
where $r$ represents $M_1 = 16$ equally spaced sampling points on the complex plane, and $P_i(r)$ are the ETDRK4 polynomial kernel functions. Every $\Delta t$ step, the physical space solution is collected to form the dataset:
\[
u(x,t) = \text{Re}\left[ \mathcal{F}^{-1}\{v(k,t)\} \right].
\]
To evaluate the model’s performance under chaotic systems with potential real-world noise, we introduce a noise distribution to the original dataset. The Gaussian distribution is generally considered to realistically reflect real-world noise interference \cite{gardner1989statistical}, so we add a standard normal distribution $Noise \sim N(0,1)$, with the hyperparameter $I_{\text{noise}}$ controlling the noise impact. The noise-affected dataset is then:
\[
u(x,t)_{\text{noisy}} = u(x,t) + \text{Noise} \times I_{\text{noise}}.
\]
Unless otherwise specified, we set the observation length $M=20$ and prediction length $L=20$.

\subsection{Without Noise}

\begin{figure}[htbp]
  \centering
  \includegraphics[width=5in]{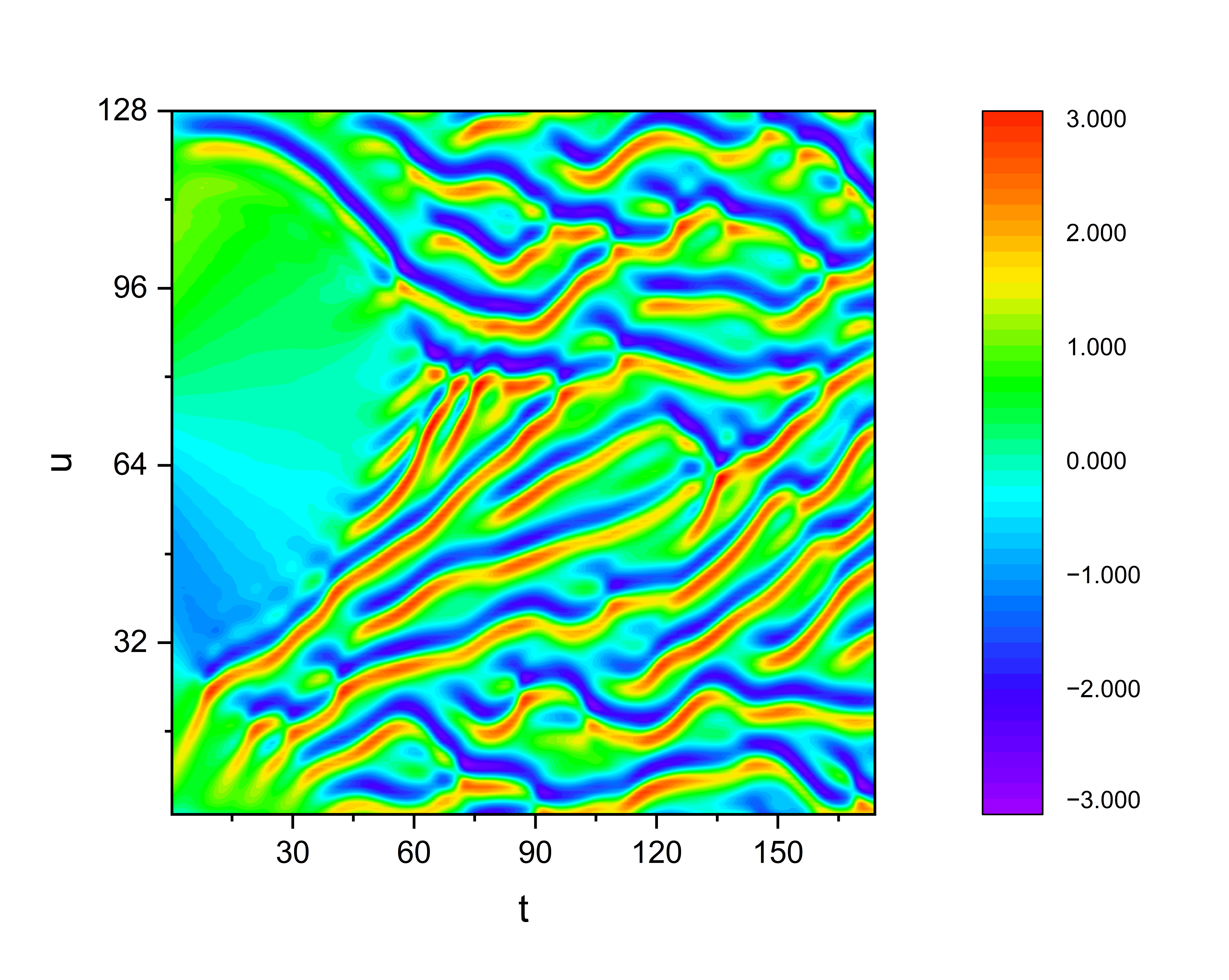} 
  \caption{Numerical solution of the equation}  
  \label{fig:6}  
\end{figure}

\begin{figure}[htbp]
  \centering
  \includegraphics[width=5in]{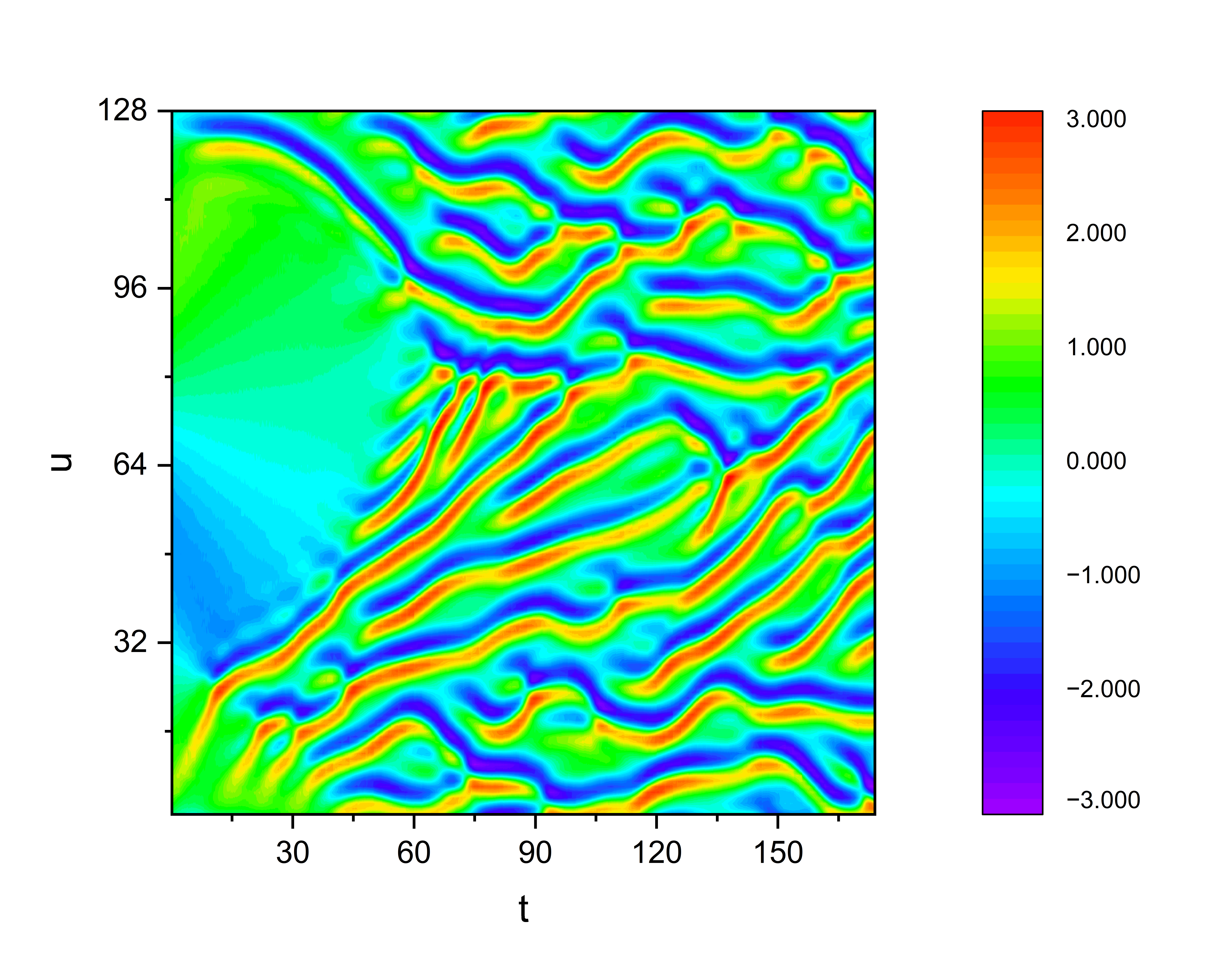} 
  \caption{Prediction results for parameters $N=128$, $t \in [0,180]$, $\text{Noise}=0$, $L=20$, $M=20$, with RMSE of 0.47314}  
  \label{fig:7}  
\end{figure}

\begin{figure}[htbp]
  \centering
  \includegraphics[width=5in]{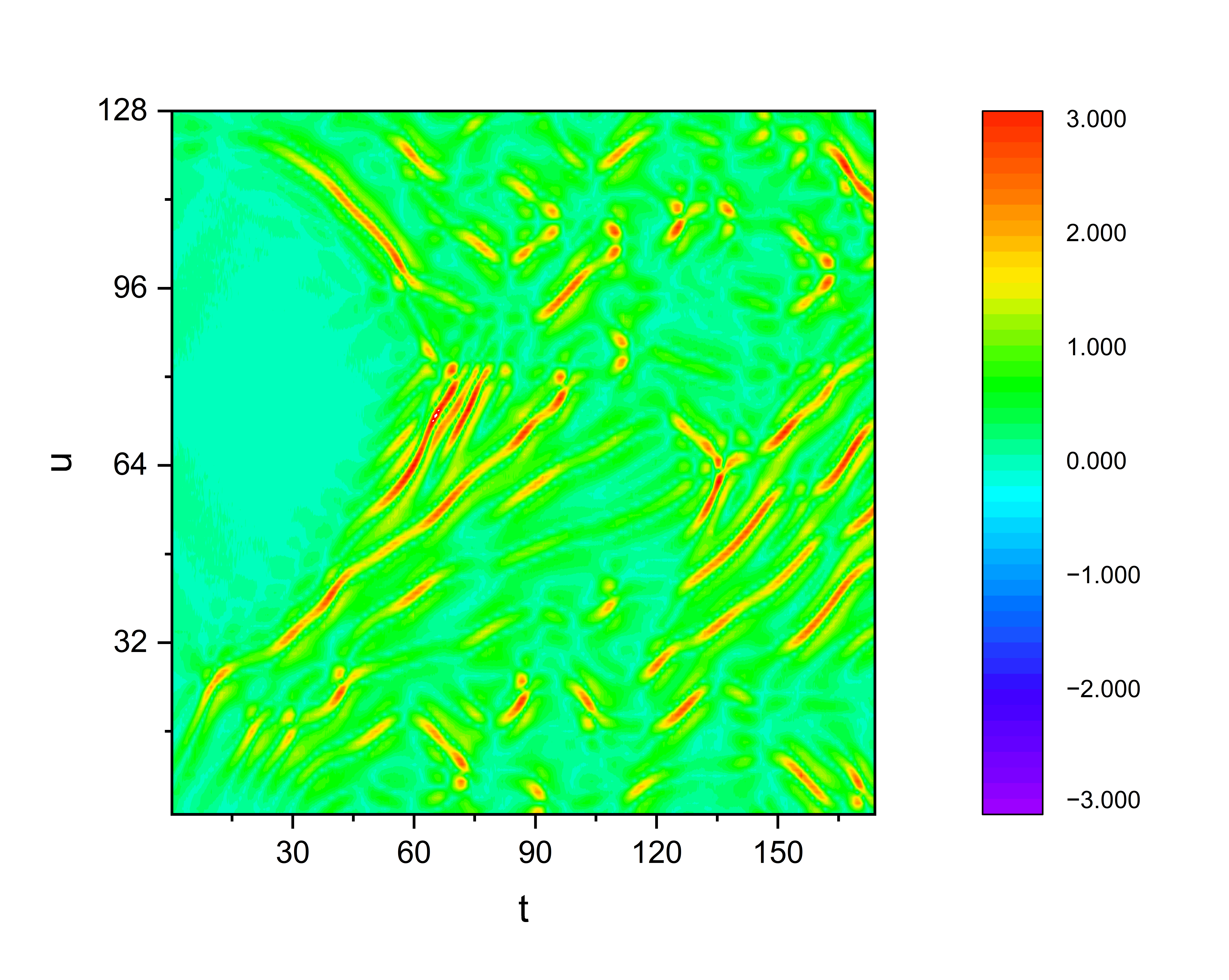} 
  \caption{Absolute error between predictions and numerical solution}  
  \label{fig:8}  
\end{figure}

To observe the model’s performance under strongly nonlinear data in Kuramoto-Sivashinsky equation, we consider its performance over the time period $t \in [0,180]$. \autoref{fig:6}, \autoref{fig:7} and \autoref{fig:8} sequentially show the numerical solution of the equation, AFD-STA predictions, and absolute error for $N=128$, $t \in [0,180]$. Overall, AFD-STA maintains accurate predictions even when the equation’s solution undergoes significant changes and the training and prediction data are in a 1:1 ratio. This indicates that AFD-STA has learned the dynamical characteristics of the nonlinear system.

The number of equally spaced points $N$ significantly affects the model’s performance. Generally, a larger number of equally spaced points better describes the system’s overall characteristics and provides more information. \autoref{Table 1} shows the results under the influence of different numbers of sampling points. We find that as the sampling interval increases, AFD-STA’s performance shows an upward trend, with the best performance at $N=128$ and stable performance at $N=256$.

\begin{table}[htbp]

\centering
\large
\begin{tabular}{c|c}
\hline
\textbf{N} & \textbf{RMSE} \\
\hline
64  & 0.63437 \\
128 & 0.47314 \\
256 & 0.49268 \\
\hline
\end{tabular}
\caption{Effect of different number of sampling points N on RMSE}
\label{Table 1}
\end{table}

\subsection{With Noise}

The previous section presented experiments without noise. However, in the real world, noise is often unavoidable due to limitations in observation techniques or uncontrollable factors. As shown in \autoref{Section 4.1}, we simulate real-world noise using a standard Gaussian distribution and control the noise level with a hyperparameter.

\begin{figure}[htbp]
  \centering
  \includegraphics[width=5in]{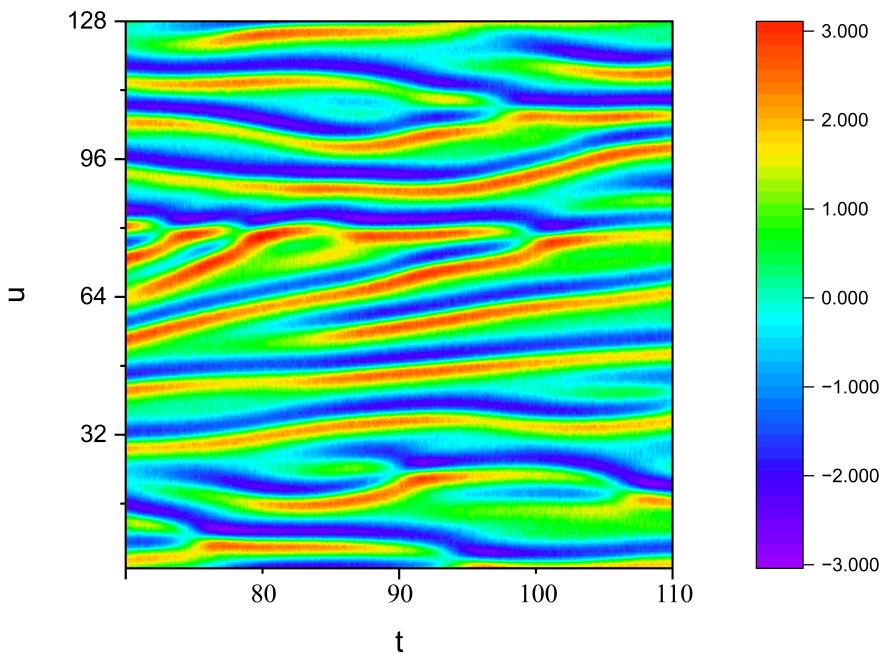} 
  \caption{Numerical solution of the equation under noise interference}  
  \label{fig:9}  
\end{figure}

\begin{figure}[htbp]
  \centering
  \includegraphics[width=5in]{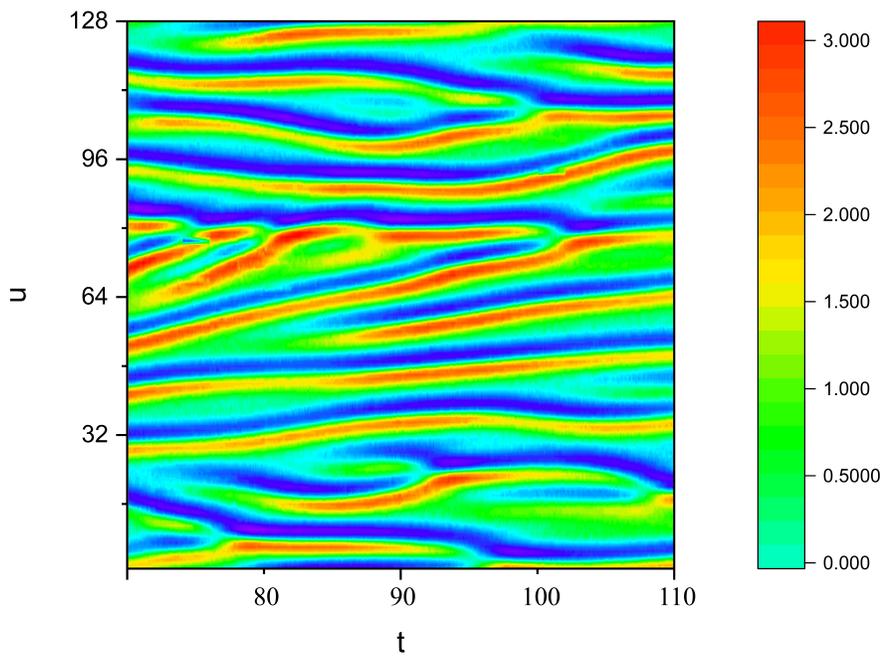} 
  \caption{Prediction results for parameters $N=128$, $t \in [70,110]$, $\text{Noise}=0.1$, $L=20$, $M=20$, with RMSE of 0.55742}  
  \label{fig:10}  
\end{figure}

\begin{figure}[htbp]
  \centering
  \includegraphics[width=5in]{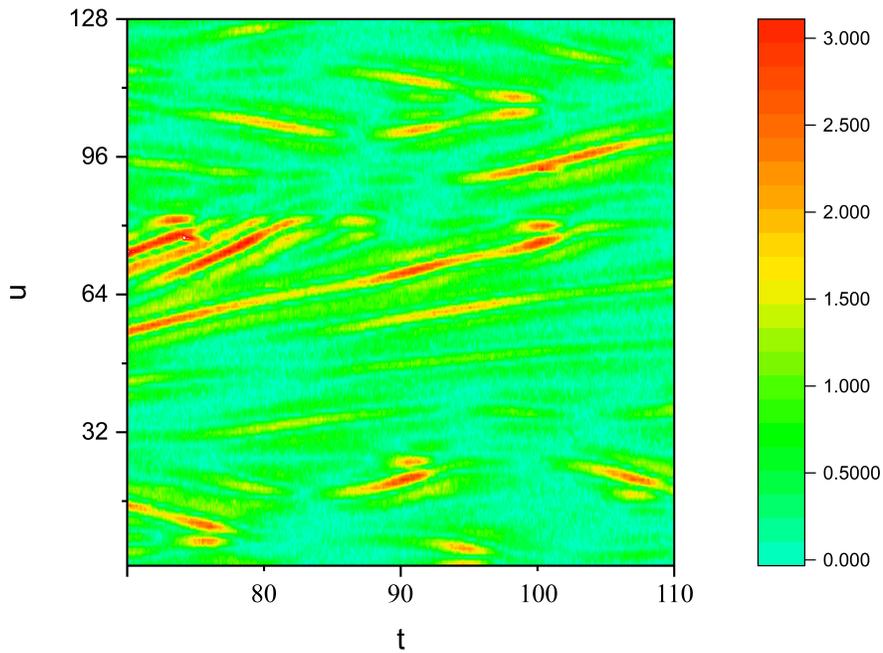} 
  \caption{Absolute error between predictions and numerical solution under noise interference}  
  \label{fig:11}  
\end{figure}

With parameters $N=128$, $\text{noise}=0.10$, we evaluate the model’s performance in a time period with significant equation changes, $t \in (70,110)$. \autoref{fig:9}, \autoref{fig:10}, and \autoref{fig:11} sequentially show the numerical solution of the equation, model predictions, and absolute error under these parameters. We find that even in a strongly nonlinear variation interval with noise interference, our model maintains accurate predictions, with larger errors only at a few individual points, demonstrating the model’s robust performance.

To test the impact of noise on our model, we conducted comparative experiments. In typical data application scenarios, the greater the noise interference, the more difficult it is to accurately predict the data. As shown in \autoref{Table 2}, as the noise parameter gradually increases, the model’s prediction performance shows a declining trend.

\begin{table}[h!]
\centering
\large
\begin{tabular}{c|c}
\hline
\textbf{Noise} & \textbf{RMSE} \\
\hline
0    & 0.53383 \\
0.1  & 0.55742 \\
0.15 & 0.58342 \\
0.2  & 0.61992 \\
\hline
\end{tabular}
\caption{Effect of different noise parameters on RMSE}
\label{Table 2}
\end{table}

\subsection{Ablation Experiment}

To verify whether each component contributes to the model and to what extent, we designed ablation experiments by sequentially removing each component. With parameters $N=128$, $L=20$, $M=20$, $\text{Noise}=0.2$, single-step predictions are performed in the region $t \in (120,122)$. Our model consists of four components: Adap-EWMA, Spatiotemporal Attention, Fusion, and DynaFC6. In each experiment, we fix the other three components, and if removing the current component reduces the model’s prediction accuracy, it indicates that the component has a positive effect on the model. Since FC-6 handles the task of dimension adjustment, removing it directly prevents obtaining the corresponding attractor, so we replace it with a simple linear layer. As shown in \autoref{Table 3}, each component of the model contributes positively to the final performance, with the Spatiotemporal Attention module contributing the most.

\begin{table}[h!]

\centering
\renewcommand{\arraystretch}{1.5}
{\large
\begin{tabular}{l|ccccc}
\hline
  \multicolumn{1}{c|}{\textbf{Component}} & \multicolumn{5}{c}{\textbf{Choice}} \\ 
\hline
Adpa-EWMA   & \checkmark &        & \checkmark & \checkmark & \checkmark \\
Spatiotemporal Attention   & \checkmark & \checkmark &        & \checkmark & \checkmark \\
Fusion & \checkmark & \checkmark & \checkmark &        & \checkmark \\
DynaFC-6        & \checkmark &\checkmark         &\checkmark         & \checkmark &  \\
\hline
RMSE        & 0.38286 & 0.40698 & 0.41729 & 0.39357 & 0.39845 \\
\bottomrule
\end{tabular}}
\caption{Ablation study of different components}
\label{Table 3}
\end{table}

The experiments show that removing the Spatiotemporal Attention mechanism leads to the largest performance loss, with RMSE increasing by 9.0\% ($\Delta=0.034$) relative to the baseline. The absence of the gated fusion module increases RMSE by 6.8\%, while removing the adaptive EWMA and replacing the FC-6 layer cause performance degradations of 6.3\% and 4.0\%, respectively. Notably, the complete model (Baseline) has an RMSE of 0.383, significantly lower than the worst value (0.417) when individual components are removed, with a performance improvement ($\Delta=0.034$) exceeding 50\% of the sum of individual component contributions, indicating a positive synergistic effect between modules.

\subsection{Comparisons Experiment}

Our model demonstrates accurate prediction performance in both noise-free and noisy datasets. To evaluate our model’s performance against other outstanding models, we select parameters $N=128$, $L=20$, $M=20$, $\text{Noise}=0.2$, $t_{\text{gap}}=0.1$, and perform single-step predictions in the region $t \in (120,122)$ in Kuramoto-Sivashinsky equation. As shown in \autoref{Table 4}, our model exhibits a significant advantage over other models across various metrics.

On the other hand, similar to the Kuramoto-Sivashinsky equation, the Brusselator equation and the Swift-Hohenberg equation are widely used nonlinear reaction-diffusion models, encompassing rich dynamical behaviors. To test whether AFD-STA outperforms other models in other nonlinear systems, we conduct experiments on the Brusselator equation and the Swift-Hohenberg equation.As shown in \autoref{Table 5} and \autoref{Table 6}, we find that AFD-STA performs the best across various metrics. LSTM \cite{hochreiter1997long} and XGBoost \cite{chen2016xgboost} are widely used in time series tasks and achieved good results in this prediction task.

Phase space reconstruction techniques, by constructing the original attractor and delayed attractor, integrate temporal variation relationships and spatial geometric features. The $i$-th row of the delayed attractor $Z(t_i) = [x_k(t_i), x_k(t_{i+1}), \ldots, x_k(t_{i+M-1})] \in \mathbb{R}^M$ can be regarded as the result of predicting $i-1$ steps based on the original attractor $O$. Thus, we believe that $L+1$ predictions enable the model to incorporate more original information, uncovering relationships in time series changes. Additionally, since the $L+1$ predictions are considered relatively independent, and the final result evaluation uses each predicted time step, this reduces model overfitting while enhancing robustness, minimizing unstable performance due to noise interference in specific cases.

\begin{table}[h!]

\centering
\renewcommand{\arraystretch}{1.25}
{\large

\begin{tabular}{l|cccc}
\hline
\textbf{Model} & \textbf{RMSE} & \textbf{MAE} & \textbf{SAMPE} & \textbf{MAD} \\
\hline
AFD-STA      & 0.46846 & 0.41038 &  57.89778 & 0.18107 \\
\hline
LSTM     & 0.50263 & 0.43640 &  58.83854 & 0.21419 \\
\hline
XGBoost  & 0.49507 & 0.43183 &  60.00500 & 0.19873 \\
\hline
CNN      & 0.54222 & 0.47550 &  64.71703 & 0.21977 \\
\hline
DNN      & 0.65152 & 0.57656 &  65.44706 & 0.24855 \\
\hline
Dlinear  & 1.01890 & 0.93453 & 117.15215 & 0.29549 \\
\bottomrule
\end{tabular}}
\caption{Single-step prediction of Kuramoto-Sivashinsky equation with parameters $N=128$, $t_{\text{start}}=120$, $L=20$, $M=20$, $t_{\text{gap}}=0.1$, $\text{Noise}=0.2$.}
\label{Table 4}
\end{table}

\begin{table}[h!]

\centering
\renewcommand{\arraystretch}{1.25}
{\large

\begin{tabular}{l|cccc}
\hline
\textbf{Model} & \textbf{RMSE} & \textbf{MAE} & \textbf{SAMPE} & \textbf{MAD} \\
\hline
AFD-STA      & 0.23778 & 0.19341 & 50.83187 & 0.15899 \\
\hline
LSTM     & 0.27289 & 0.22177 & 52.06842 & 0.19161 \\
\hline
XGBoost  & 0.26578 & 0.21511 & 55.47640 & 0.17562 \\
\hline
CNN      & 0.28124 & 0.22733 & 58.94235 & 0.17178 \\
\hline
DNN      & 0.38313 & 0.33157 & 85.43416 & 0.15879 \\
\hline
Dlinear  & 0.48980 & 0.43149 & 114.65781 & 0.18528 \\
\hline
\end{tabular}}
\caption{Single-step prediction of Brusselator equation with parameters $N=128$, $t_{\text{start}}=0.1$, $L=20$, $M=20$, $t_{\text{gap}}=0.00001$, $\text{Noise}=0.2$.}
\label{Table 5}
\end{table}

\begin{table}[h!]

\centering
\renewcommand{\arraystretch}{1.25}
{\large

\begin{tabular}{l|cccc}
\hline
\textbf{Model} & \textbf{RMSE} & \textbf{MAE} & \textbf{SAMPE} & \textbf{MAD} \\
\hline
AFD-STA      & 0.23695 & 0.19338 & 81.76329 & 0.15645 \\
\hline
LSTM     & 0.27888 & 0.22513 & 90.01469 & 0.18522 \\
\hline
XGBoost  & 0.27282 & 0.22116 & 88.75923 & 0.17777 \\
\hline
CNN      & 0.26262 & 0.21159 & 89.92485 & 0.16986 \\
\hline
DNN      & 0.27962 & 0.22765 & 94.41591 & 0.18603 \\
\hline
Dlinear  & 0.31240 & 0.25400 & 106.24605 & 0.18757 \\
\hline
\end{tabular}}
\caption{Single-step prediction of Swift-Hohenberg equation with parameters $N=128$, $t_{\text{start}}=5$, $L=20$, $M=20$, $t_{\text{gap}}=0.001$, $\text{Noise}=0.2$.}
\label{Table 6}
\end{table}

\section{Conclusion}
The proposed AFD-STA framework breaks through traditional methods' dependence on initial conditions, directly learning intrinsic dynamical characteristics of spatiotemporal chaotic systems from observational data through phase space reconstruction for prediction. This method only requires training data equivalent to the prediction duration (1:1 ratio), achieving high-precision predictions in canonical systems including Kuramoto-Sivashinsky, Brusselator, and Swift-Hohenberg. Compared with other models, AFD-STA maintains stable performance under Gaussian noise interference with $\sigma \leq 0.2$.

The current AFD-STA architecture still has limitations in long-term predictions of attractor mutations, primarily stemming from the time-varying characteristics of Lyapunov exponent spectra in infinite-dimensional spatiotemporal chaotic systems. Our AFD-STA network can only focus on situations with stable attractor properties in the short term, and cannot predict attractor mutations during long-term predictions. This limitation may originate from insufficient attention to the dynamic evolution trends of attractors. Future work plans to introduce the Brox variational optical flow method~\cite{brox2010large}, capturing attractor deformation trends by quantifying velocity field changes between consecutive time steps, thereby constructing an optical flow-spatiotemporal joint attention mechanism to enhance modeling capabilities for dynamic evolution of phase space structures.

\section*{Acknowledgement}
This research is supported by the Shandong Natural Science Foundation (ZR2024QD287).

\bibliographystyle{unsrt}  
\bibliography{AFD-STA}

\end{document}